\documentclass[letterpaper,twocolumn,fleqn]{article} 
\usepackage{lmodern}
\usepackage{ist}
\usepackage{graphicx}
\usepackage{float}

\title{Color in Visual-Language Models: CLIP deficiencies}

\author{Guillem Arias, Ramon Baldrich, Maria Vanrell \\
Computer Vision Center / Universitat Autònoma de Barcelona}

\date{} 

\begin{document} 

\maketitle 

\thispagestyle{empty} 


\begin{abstract}
This work explores how color is encoded in CLIP (Contrastive Language-Image Pre-training) which is currently the most influential VML (Visual Language model) in Artificial Intelligence. After performing different experiments on synthetic datasets created for this task, we conclude that CLIP is able to attribute correct color labels to colored visual stimulus, but, we come across two main deficiencies: (a) a clear bias on achromatic stimuli that are poorly related to the color concept, thus white, gray and black are rarely assigned as color labels; and (b) the tendency to prioritize text over other visual information. Here we prove it is highly significant in color labelling through an exhaustive Stroop-effect test. With the aim to find the causes of these color deficiencies, we analyse the internal representation at the neuron level. We conclude that CLIP presents an important amount of neurons selective to text, specially in deepest layers of the network, and a smaller amount of multi-modal color neurons which could be the key of understanding the concept of color properly. Our investigation underscores the necessity of refining color representation mechanisms in neural networks to foster a more comprehensive comprehension of colors as humans understand them, thereby advancing the efficacy and versatility of multimodal models like CLIP in real-world scenarios.
\end{abstract}

\section{1. Introduction}
\label{sec:intro}

In the last decade artificial intelligence (AI) has significantly progressed in the construction of deep trained models able to solve vision and language problems with a significant efficiency. One particular achievement has been the construction of multimodal models, namely Visual-Language models (VLMs) with the ability to associate textual and visual descriptions to seek different tasks \cite{SurveyVLM2024}. The main advantage of these models over traditional CNNs is their ability to correlate any given text with any given image thanks to the combination of their text encoder and image encoder. One VLM worth mentioning is CLIP (\textit{Contrastive Language–Image Pre-training}), which has been the most successful due its competitive results and generalization capabilities. It was the first model trained with contrastive learning \cite{chen2020simple} and was introduced by OpenAI in 2021 \cite{CLIP2021}, who released the model to the community accelerating its impact both in research and industry applications. 

CLIP is composed by an image encoder which is a Resnet-like architecture \cite{he2016deep}, plus a text encoder based on a transformer \cite{vaswani2017attention}, finally a cosine similarity between the encoded texts and the encoded images is computed as a measure of how likely a certain text represents an input image. One of the main advantages of CLIP is how it excels at zero-shot learning, allowing it to perform a task without needing task-specific fine-tuning thanks to its generalization capability and  leveraging its understanding of a wide range of visual and textual concepts. This is achieved through an initial pre-training of the visual encoder with ImageNet dataset \cite{imagenet_cvpr09}, followed by a subsequent  training using 400 million of image-text pairs that were collected from the internet. These pairs consist of images with their corresponding captions or descriptions. Thanks to this training, CLIP learns to associate images with their corresponding text descriptions and distinguishing them from unrelated image-text pairs using a contrastive loss function.

CLIP is specially interesting in color categorization because its ability to understand text anso improves its color categorization as shown in \cite{Akbarinia2024colour}.This improvement could be due to one of the most interesting properties that emerge from this training: multi-modal neurons\cite{goh2021multimodal},\cite{Salin2022VisionLanguage}. Multi-modal neurons are units with a strong activation for a certain concept regardless of its representation (text, realistic image, drawing ...). This is reminiscent of a similar phenomenon in some human neurons, which fire in response to images, whether they are photos, drawings, or even words of the same concept \cite{Quiroga2005}, making this model specially interesting to study for its parallelism with the human brain.

\begin{figure}[t]
\centering
\includegraphics[width=\linewidth]{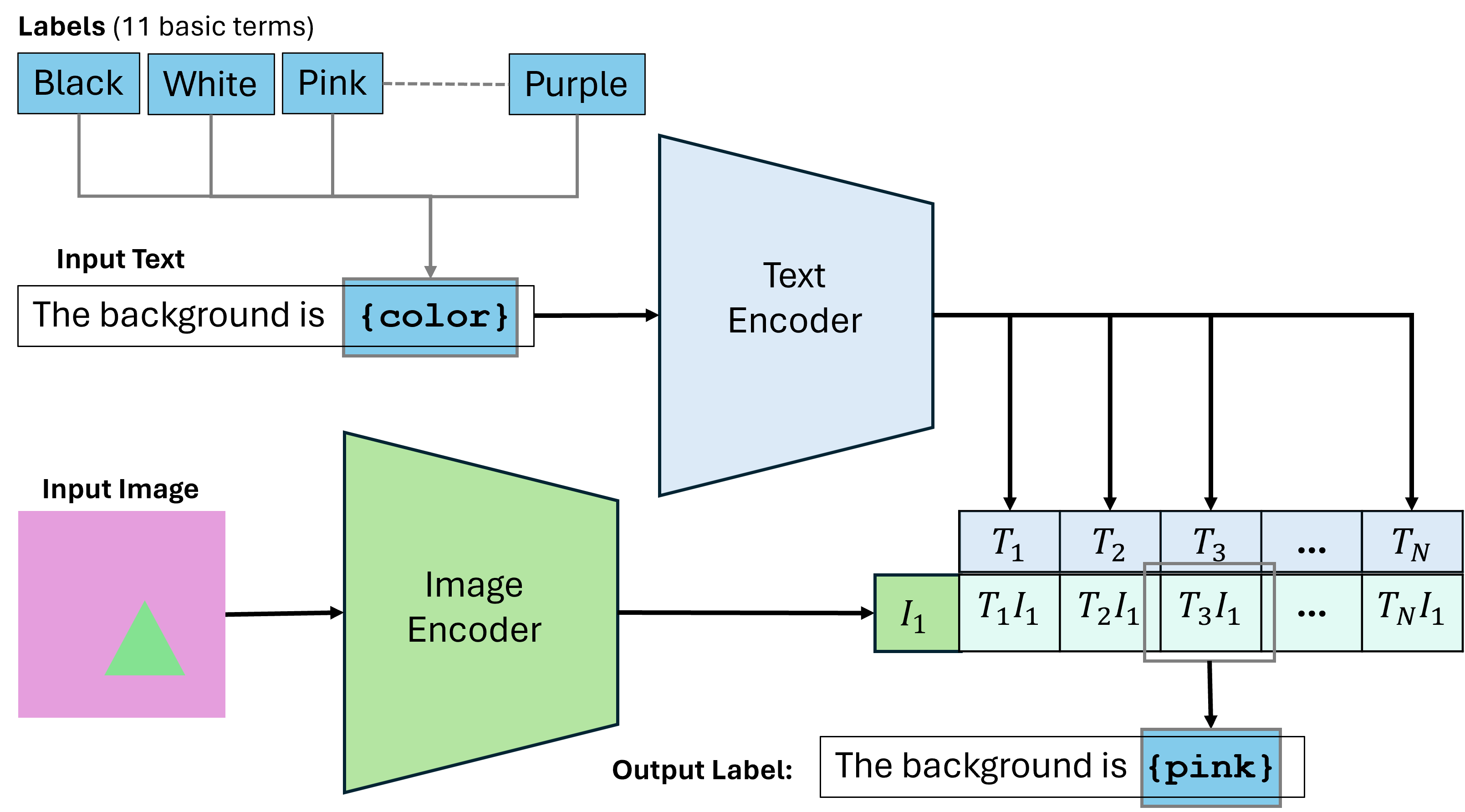}
\caption{CLIP architecture set for a color naming task. The input image in the visual encoder is contrasted with several color labels within the input text. The output is the label that maximizes the visual and text embedding (Example: Input Text is "The background is \texttt{\{ color \}}" and Input Image is a green triangle with a pink background).}\label{fig:CLIP-scheme}
\end{figure}

As the rest of Deep Neural Networks engines, CLIP presents a black-box nature and lacks a clear explanation about how knowledge is embedded in both encoders.  Understanding multi-modal neurons can sheer light on how CLIP works, by analyzing what stimulus activates them. On this topic, \cite{goh2021multimodal} research proves the concept of multi-modal neurons within CLIP, which respond to specific concepts across both text and image domains and 
 MultiViz framework \cite{liang2023multiviz} provides a comprehensive approach to analyzing and understanding the internal mechanics of multi-modal models like CLIP by identifying the importance of individual inputs in the overall prediction process, examining the relationships and dependencies between multiple inputs and interpreting the contribution of every features to the output.

Although previous works have delved into the categorization and understanding of multi-modal neurons, their main focus is on the deeper layers of the model, where higher level concepts are formed, but there is not much research about how low-level features such as color are learned in this models and if there is presence of multi-modal neurons in lower layers of CLIP.

The aim of this work is to explore how color is learned in CLIP. We first explore if color is understood as an object attribute by asking questions on a basic color/object dataset. Then we explore the capability of CLIP to recognize and read color with a Stroop Test dataset. Previous works \cite{goh2021multimodal} have stated that CLIP, like humans present the Stroop effect, we analyse this fact in more depth, we conclude that CLIP has some problems in recognizing achromatic colors as visual color attributes and we propose a new index and a wider set of experiments to identify why these color problems emerge in CLIP. Finally we analyse the internal representation of color at the neuron level. We compute the distribution of color selective neurons showing a similar distributions as object recognition models \cite{rafegas2018}. We also explore what neurons are activated in the Stroop task and find color multi-modal neurons emerging in earlier layers of the neural network.



\section{2. Color predictions on basic images}

In this section we aim to perform preliminary experiments to explore how CLIP associates color labels to a particular image. To test this ability, we have created a dataset of images containing an homogeneous color background and one basic colored shape. To generate these images, we have used  8 basic shapes (triangle, square, circle, amongst others) and 11 representative universal colors which are colors with a common color term in most developed languages \cite{berlin1969basic} with fixed RGB values. The dataset contains 500 images for each possible combination with different rotations, positions and scale of the shapes,totalling 440,000 images (8 shapes x 11 background colors x 10 object colors x 500 samples). Setting CLIP with the color labels to be associated in the input text as it is shown in figure \ref{fig:CLIP-scheme} we perform the following experiments:

\paragraph{Experiment 1.} In the first experiment we evaluate how CLIP associates a color term to an image in global, without asking for any specific part of the input image. The input text is just the color label. This allow us to determine if CLIP presents any prior bias towards any specific color, or whether it is inclined to answer with the most predominant color of the image, no matter if it is the object or the background.  The results are shown in Table \ref{tab:experiment1}. We can see that CLIP is inclined to assign color labels which are Chromatic. When the background is achromatic and the object not, it predicts the color of the object ($91.08\%$), and when the object is achromatic and background not, then it returns the color of the background ($99.74\%$). However, when both, object and background are the same, it predicts the color of the background, $63.80\%$ in Achromatic combinations and $78.62\%$ in Chromatic combinations.

\begin{table}[H]
\centering
\begin{tabular}{|c|c|c|c|} \hline
\multicolumn{4}{|c|}{\small{Input Text:  \{\_\_\_\_\_\_\}  \tiny{\textit{(One color label for the full image)}}}} \\ \hline
\multicolumn{2}{|c|}{\small{CLIP prediction }} &  \multicolumn{2}{|c|}{\textit{\small{Input Object}}} \\ \cline{3-4}
\multicolumn{2}{|c|}{\scriptsize{\textbf{Background / Object}}} &
\textit{\small{Achromatic}}  &  \textit{\small{Chromatic}}  \\\hline
\textit{\small{Input}} & \textit{\small{Achrom.}} &  \textbf{63.80\%} / 35.96\% &   6.28\% / \textbf{91.08\%} \\ \cline{2-4}
\textit{\small{Backg.}} & \textit{\small{Chrom.}} & \textbf{99.74\%} / 0.19\% & \textbf{78.62\%} / 20.42\% \\ \hline
\end{tabular} \vspace{0.05cm}
\caption{Experiment 1. CLIP global color prediction to the Input Text in top row. Ratios for Background Color / Object color assignment depending of Input Image Type. Input Images divided in 4 groups depending on the chromaticity of Object and Background. Remaining ratios are incorrect color assignments.}\label{tab:experiment1} \vspace{0.4cm}
\end{table}
\paragraph{Experiment 2.} In a second experiment, we evaluate if CLIP can attribute color to a specific part of the image.  We ask CLIP to associate a color label to the image object or to the background accordingly with the Input Text indicated in Table \ref{tab:experiment2}. In this way we can test if CLIP properly predicts the semantic information embedded in the input question regarding color attribution.  The results are shown in Table \ref{tab:experiment2}. When we ask for the color of the Object, CLIP predicts the correct color with 73\%, 92\% and 83\%, but when the object is achromatic the performance decreases to 0.19\%, and gives the color of the background. When we ask for the color of the Background, CLIP predicts the correct color with 70\%, 99\% and 81\%, but when the background is achromatic and the object chromatic the performance decreases to 5\%, and gives the color of the object. This brings to the same conclusion as in experiment 1, CLIP does not attribute the color word to achromatic parts of the image in presence of a Chromatic color, neither the object nor the background, but can properly link colors to the concept of object and background when both stimulus are of the same mode (Chromatic or Achromatic).

\begin{table}[H]
\centering

\begin{tabular}{|c|c|c|c|} \hline
\multicolumn{4}{|c|}{\small{Input Text: The color of the object is \{\_\_\_\_\_\_\} }} \\ \hline
\multicolumn{2}{|c|}{\small{CLIP prediction }} &  \multicolumn{2}{|c|}{\textit{\small{Input Object}}} \\ \cline{3-4}
\multicolumn{2}{|c|}{\scriptsize{\textbf{Background / Object}}} &
\textit{\small{Achromatic}}  &  \textit{\small{Chromatic}}  \\\hline
\textit{\small{Input}} & \textit{\small{Achrom.}} &  24.88\% / \textbf{73.69\%} &   4.61\% / \textbf{92.85\%} \\ \cline{2-4}
\textit{\small{Backg.}} & \textit{\small{Chrom.}} & \textbf{99.83\%} / 0.15\% & 16.50\% / \textbf{83.41\%} \\ \hline
\multicolumn{4}{c}{} \\ \hline
\multicolumn{4}{|c|}{\small{Input Text: The color
of the background is \{\_\_\_\_\_\_\} }} \\ \hline

\multicolumn{2}{|c|}{\small{CLIP prediction }} &  \multicolumn{2}{|c|}{\textit{\small{Input Object}}} \\ \cline{3-4}
\multicolumn{2}{|c|}{\scriptsize{\textbf{Background / Object}}} &
\textit{\small{Achromatic}}  &  \textit{\small{Chromatic}}  \\\hline
\textit{\small{Input}} & \textit{\small{Achrom.}} &  \textbf{70.88\%} / 28.74\% &   5.46\% / \textbf{93.16\%} \\ \cline{2-4}
\textit{\small{Backg.}} & \textit{\small{Chrom.}} & \textbf{99.95\%} / 0.02\% & \textbf{81.34\%} / 18.61\% \\ \hline

\end{tabular} \vspace{0.05cm}
\caption{Experiment 2. CLIP prediction of color attribution to an image part for two different Input Text on the top row of each table. Ratios of ( Background Color / Object Color ) assignment depending of Input Image Type. Input Images divided in 4 groups depending on the chromaticity of Object and  Background. Remaining ratios are incorrect color assignments.}\label{tab:experiment2} \vspace{0.4cm}
\end{table}

\section{3. Color predictions on text}

Once we have explored CLIP performance in color assignment to basic images,  we will explore how color behaves in  colored text. To this end, we have set the classical Stroop test \cite{stroop1935} where the task is to predict the color of a word in a colored font, with the particularity that the word is a color name. With this task we want to evaluate if CLIP is able to distinguish the semantics of a question that asks not to read but to perceive the color. 

\paragraph{Experiment 3.}Before performing the Stroop test on CLIP, we evaluated different options of input text for this task to see if CLIP could understand the question asked properly. In table \ref{tab:QuestEvaluation} we show the distribution of answers for 5 different Input Text sentences. We can see that no matter the question asked about the color of the text, the answers present a clear bias towards answering the color name. Therefore, CLIP seems to be clearly more inclined to read than to use any other visual information, as already was proved in a previous work\cite{goh2021multimodal}. Considering the results given in this table we decided to use the top question that gives the lower ratio in failing in the Stroop test task (59.53\%), that is the best in giving the color of the font (although 2.23\% is not significant at all), which was better understood than the question in the bottom row which was the one used in a similar experiment performed in \cite{goh2021multimodal}.

\begin{table}[H]
\hspace{-0.25cm}\centering
\begin{tabular}{|l|c|c|c|c|} \hline
  & \scriptsize{Backg.} & \scriptsize{Font} & \scriptsize{Written} & \scriptsize{None} \\
\scriptsize{Input Text} & \scriptsize{Color} & \scriptsize{Color} & \scriptsize{Color} & \tiny{in Input} \\ \hline
\scriptsize{The word is written in}  & \small{38.03\%} & \small{\textbf{2.35\%}} & \small{59.53\%} & \small{0.09\%} \\
\scriptsize{ \{\_\_\_\_\_\_\}} font &  & &  &  \\ \hline
\scriptsize{The text says \{\_\_\_\_\_\_\}} & \small{13.95\%} & \small{0.81\%} & \small{\textbf{85.16\%}} & \small{0.08\%} \\ \hline
\scriptsize{The color of the  }&  \small{\textbf{38.63\%}} & \small{1.69\%} & \small{59.63\%} & \small{0.05\%} \\ 
\scriptsize{ background is \{\_\_\_\_\_\_\} }&  &  &  & \\ \hline
\scriptsize{\{\_\_\_\_\_\_\}} & \small{21.78\%} & \small{1.07\%} & \small{77.07\%} & \small{0.09\%} \\ \hline
\scriptsize{My favorite word,}  & \small{36.98\%} & \small{1.98\%} & \small{61.01\%} & \small{0.04\%} \\
\scriptsize{written in the color} \{\_\_\_\_\}& &  &  &  \\\hline
\end{tabular}\vspace{0.05cm}
\caption{Experiment 3. Evaluation of questions. Ratio of color label answers for each image 
 with a different color for background, font or word.}\label{tab:QuestEvaluation} \vspace{0.4cm}
\end{table}

\label{sec:colorinclip}

In what follows we show the results of two different Stroop experiments on CLIP. To assess the performance on this task we have generated a Stroop dataset with 11 basic color names written in 10 different basic colors (excluding the color name) and for 9 different colored backgrounds (excluding the  color name and the font color) with 500 samples of each (varying the type of font, size and position) that totals on 495,000 images. The first experiment, which is similar to the original test where colored color terms appear on a white background, having to chose mainly between the written color, or the color of the font. In a second experiment, we use a larger set of images where the background is colored. This experiment adds a new color distractor to the task.

\paragraph{Experiment 4. Stroop Test on white background.}
The results of this experiment are summarized in Table \ref{tab:experiment4} . In the first two columns we show the percentage of correct answers, this is CLIP returning the color of the FONT of the input image. We can see the accuracy for the task is very low, only for 16.7\% of the images we get the correct color answer. In the third and fourth columns we give the ratio of incorrect answer, this is, CLIP assigning the color written in the text in 81.1\% over all images or answering a color in neither stimulus 2.17\%. These ratios are equally distributed for all colors except for grey, where we get a small minor rate.  In conclusion, from these results we can state that CLIP presents a strong Stroop effect, making it unable to avoid reading instead of focusing on other stimulus. 

\begin{table}[ht]
\centering
\includegraphics[width=1\linewidth]{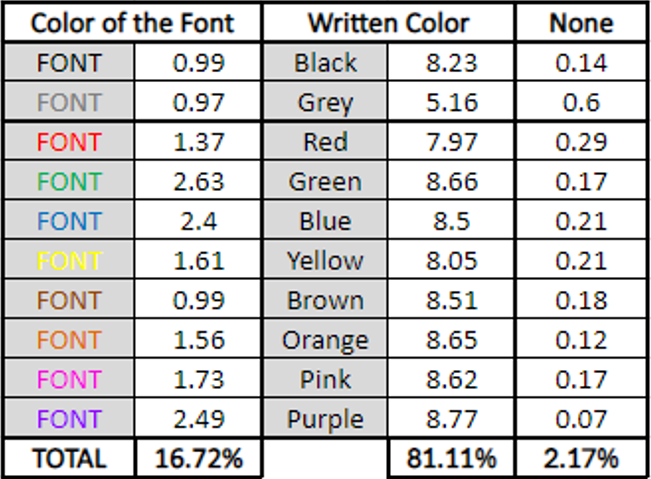}
\caption{Experiment 4. Stroop Test with White Background. \textit{(Column 2): \% of Correct Answers. (Columns 4,5): \% of Incorrect answers.}}\label{tab:experiment4} \vspace{0.4cm}
\end{table}

\paragraph{Experiment 5. Stroop Test on colored background}
The results of this experiment are shown in Table \ref{tab:experiment5}. This experiment show worst results than the previous one due the presence of a second distractor (Background). In the first two columns we show the percentage of correct answers in returning the color of the FONT of the input image. We can see that the error for this task has been notably increased with respect to the white background, only for 2.35\% of the images we get the correct color answer. In the fourth and sixth columns we give the distribution of the incorrect answers. In a 59.5\% of the cases, CLIP returns the color name, and in 38\% assigns the color of the background. It is worth mentioning that the distraction effect of the background is quite low on Achromatic colors with significant lower errors compared with the Chromatic colors, this results go in line with the results observed in the first experiments where Achromatic colors were not taken into consideration in presence of other stimuli.  This second experiment confirms the bias towards written stimuli of CLIP. 

\begin{table}[ht]
\centering
\includegraphics[width=1\linewidth]{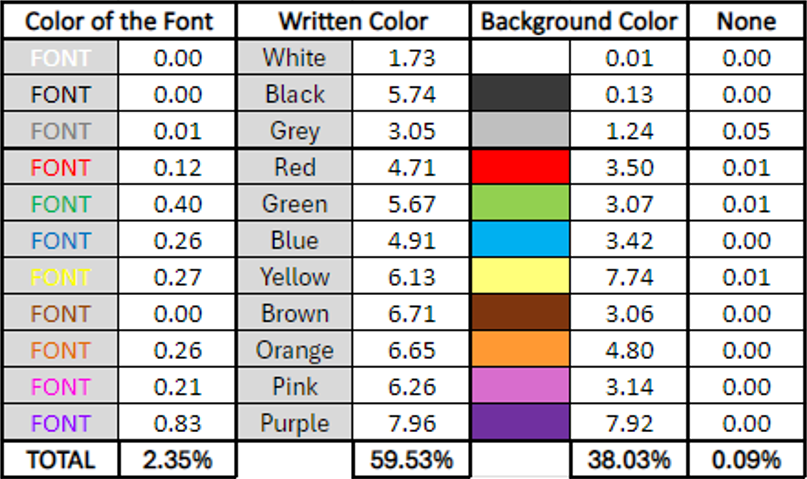}
\caption{Experiment 5. Stroop test with Colored Background. \textit{(Column 2): \% of Correct Answers. (Columns 4,6,7): \% of Incorrect answers.} }\label{tab:experiment5} \vspace{0.4cm}
\end{table}

\section{4. Color in CLIP Visual Encoder}
\subsection{Color Selectivity}
Once we concluded several deficiencies in color label assignment, in this section we explore possible causes for these drawbacks. Our exploration is done at the neuron level. Firstly, we analyse the generic Color Selectivity Index of individual neuron units as it was definded in \cite{rafegas2018}, where the Color Selectivity index is calculated by finding the 100 top scoring patches over a large dataset (Imagenet), and calculating the difference in activation by those same patches in gray scale, with a high color selectivity meaning that color is important to activate a neuron, and a low selectivity index meaning that only the shape was important to activate a neuron. Secondly, we propose an in-depth analysis of the activation of the CLIP individual neurons provoked by the Stroop dataset in CLIP by defining a Color-Label Selectivity Index, and classifying the neurons over the network based on the stimulus of their activation.

\begin{figure}
\centering
\includegraphics[width=\linewidth]{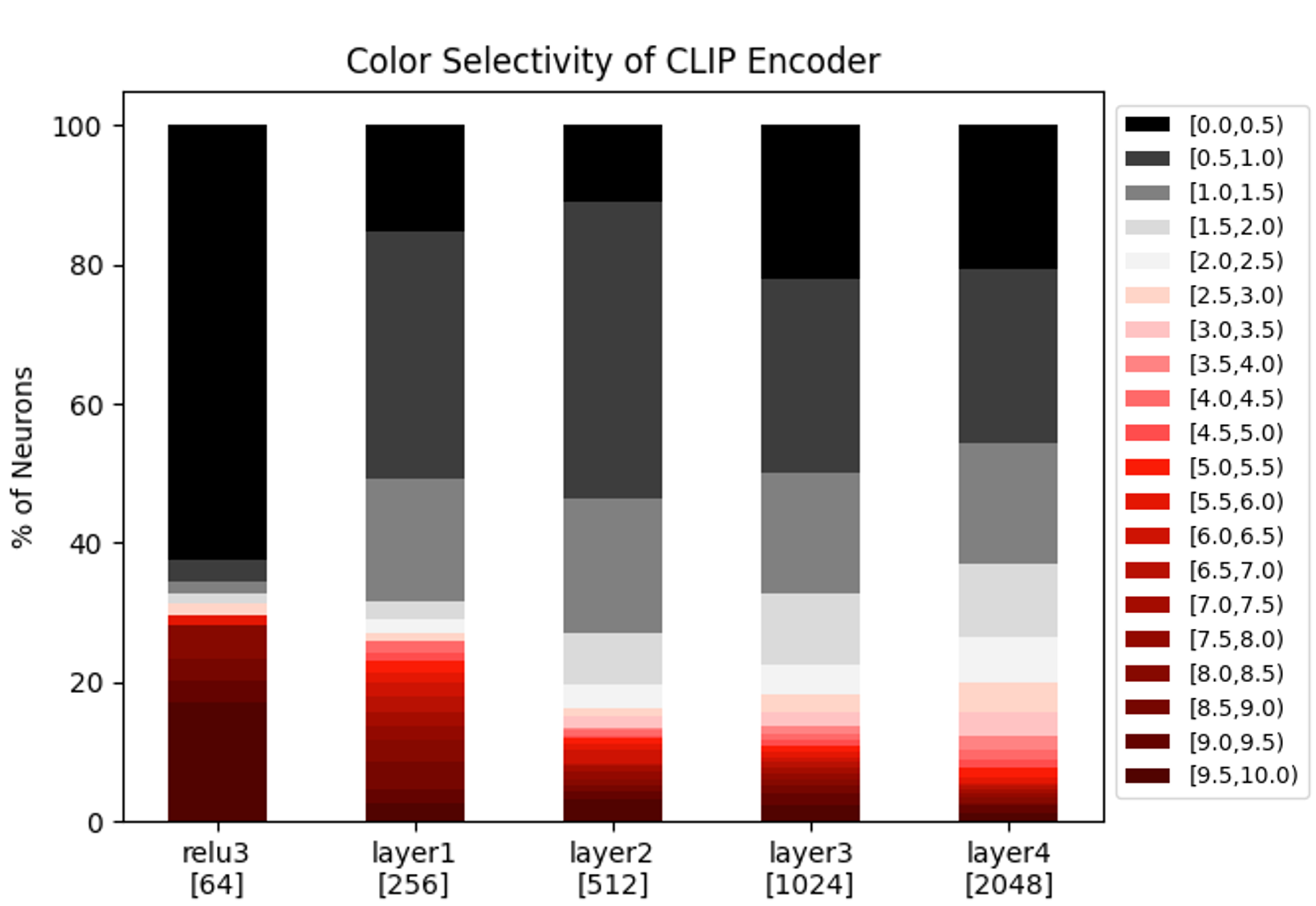}
\caption{Distribution of Color Selective Neurons in CLIP Visual Encoder Layers.}
\label{fig: cselectivity-CLIP}\vspace{-0.2cm}
\end{figure}

\begin{figure}[H]
\vspace{-0.2cm}
\centering
\includegraphics[width=\linewidth]{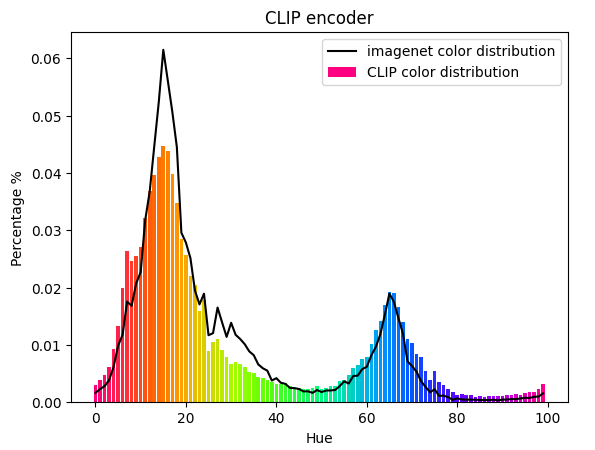}
\caption{Hue Selectivity Distribution in the Visual Encoder vs. Imagenet Distribution (Pearson's correlation coefficient R=0.965). }
\label{fig: HUEselectivity-CLIP}\vspace{-0.2cm}
\end{figure}

In figure \ref{fig: cselectivity-CLIP} we show the distribution of color selectivity indexes for the neurons of each block of convolutional layers of the CLIP Visual Encoder. Overall, we can see that the ratio of color selectivity is a bit lower than the one usually found in object recognition models (e.g. see fig.4 (a) in \cite{rafegas2018}). This could be due to the subsequent training process that CLIP goes through after the initial training on Imagenet. The training process on a large Internet dataset to acquire text understanding skills could have shifted color selective neurons to pattern selective neurons necessary to accommodate reading abilities. In figure \ref{fig: HUEselectivity-CLIP}, we show the hue distribution of the color selective neuron features we have found in CLIP. Despite the second training on a larger dataset, CLIP's hue selectivity  maintains a high correlation with the ImageNet hue distribution (Pearson's correlation coefficient of 0.965), which proves the posterior training has not reduced the overall distribution of the selectivity properties.

\subsection{Activation Analysis}
\begin{figure}[t]
\centering
\vspace{-0.2cm}
\includegraphics[width=\linewidth]{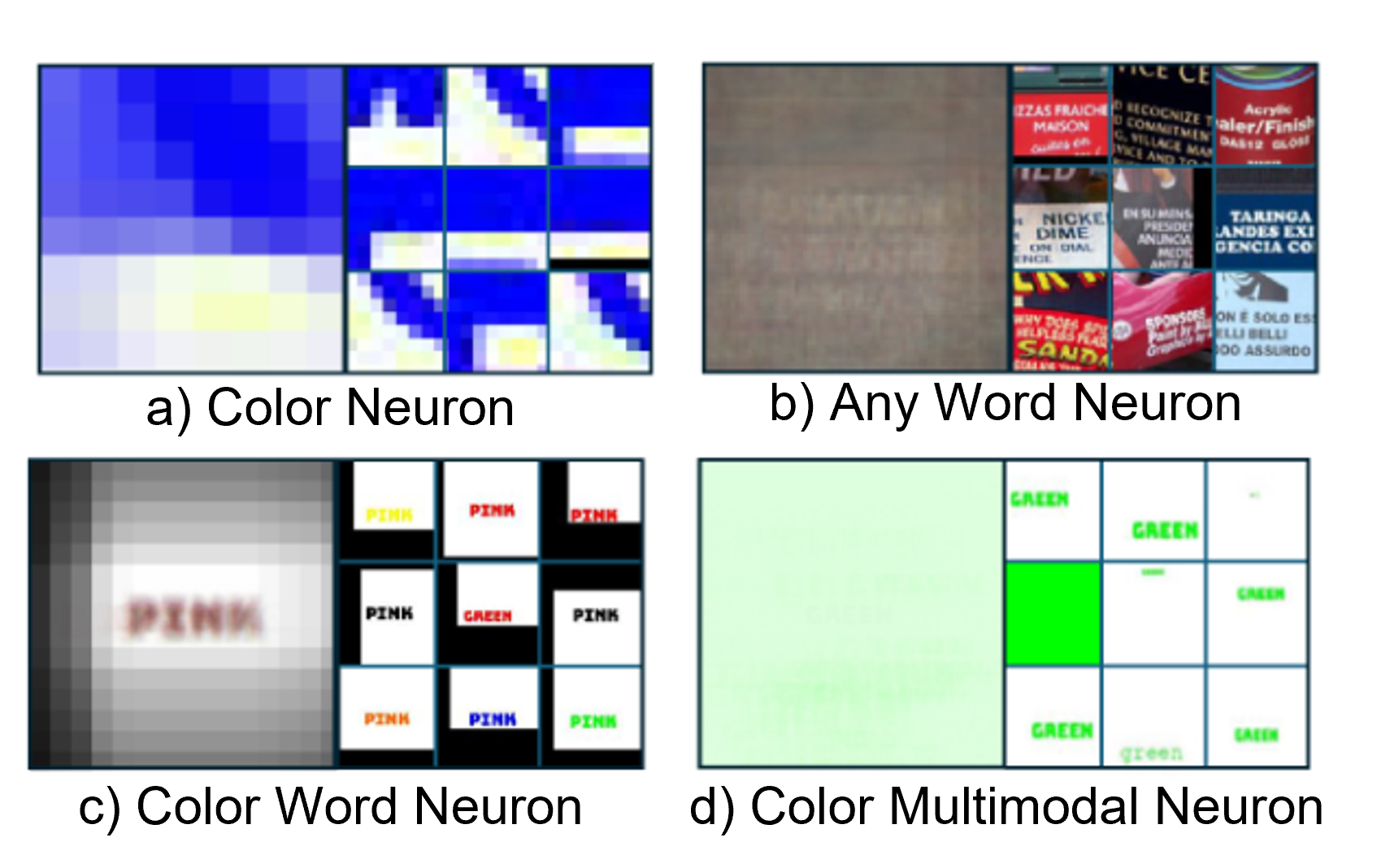}
\caption{Example of 4 types of Neurons. Left side: Neuron feature (weighted averaged of the first 100 top-scoring cropped images). Right side: 9 top-scoring cropped images from the Stroop dataset.}
\label{fig: neuron_types}\vspace{-0.2cm}
\end{figure}

In pursuing the causes of the color deficiencies found in the previous experiments, we propose a new \textit{Color-Label Selectivity Index}  inspired on the Class selectivity Index proposed by Rafegas et-al in \cite{rafegas2020understanding}. This new index, $f_c$, measures the relative frequency of each color label $c$ for a given neuron, $n_{i,L}$, and it is estimated as:

\begin{equation}
\centering
f_c\left(n^{i,L}\right) =\frac{\sum_j^{N_{c}} w_{j,i,L}}{\sum_l^{N} w_{l,i,L}}
\end{equation}

\noindent where $N_c$ refers to the number of images, among the $N$ cropped top scoring images activating this neuron that contains the specific color label $c$. Considering $n^{i,L}$ is the \textit{i-th} neuron of layer $L$, we denote its activation value as $w_{s,i,L}$, when the input image is $s$. By computing this index over all the activation's provoked by the Stroop dataset we can classify all the neurons in 5 
different categories with the following criteria:
 
\begin{description}

\item[Color:] Neuron with \textit{high Color-Label Selectivity Index for a specific color label}, independently if the label is attributed to the background or to the font, it activates for a specific color label. We can differentiate between Chromatic and Achromatic color labels. Example in Figure \ref{fig: neuron_types}.(a).

\item[Any Word:] Neuron with \textit{a high activation for any word}. It had a high activation for the Stroop Dataset as well as for images of Imagenet containing text, which means it activates for any kind of written word independently of its color or meaning. Example in Figure \ref{fig: neuron_types}.(b).
\item[Color Word:] Neuron with \textit{high Color-Label Selectivity index for a specific color word}, i.e. it activates only for one specific color name independently of the color of the font. Example in Figure \ref{fig: neuron_types}.(c).
\item[Color Multimodal:] Neuron with \textit{high Color-Label Selectivity index for a specific color word,  for the same color label of font, and for the same color label of background}, i.e. it activates for one specific color in all its modalities, it captures the concept of the color in full. Example in Figure \ref{fig: neuron_types}.(d).
\item[Not activated:] Neuron with \textit{low activation to images in Stroop dataset}, which means that those neurons are not selective to color in any of its modalities. It maximum activation in Stroop dataset does not reach 50\% of the maximum activation that this neuron achieves with the ImageNet dataset.
\end{description}

\begin{figure}[t]
\centering
\includegraphics[width=1.1\linewidth]{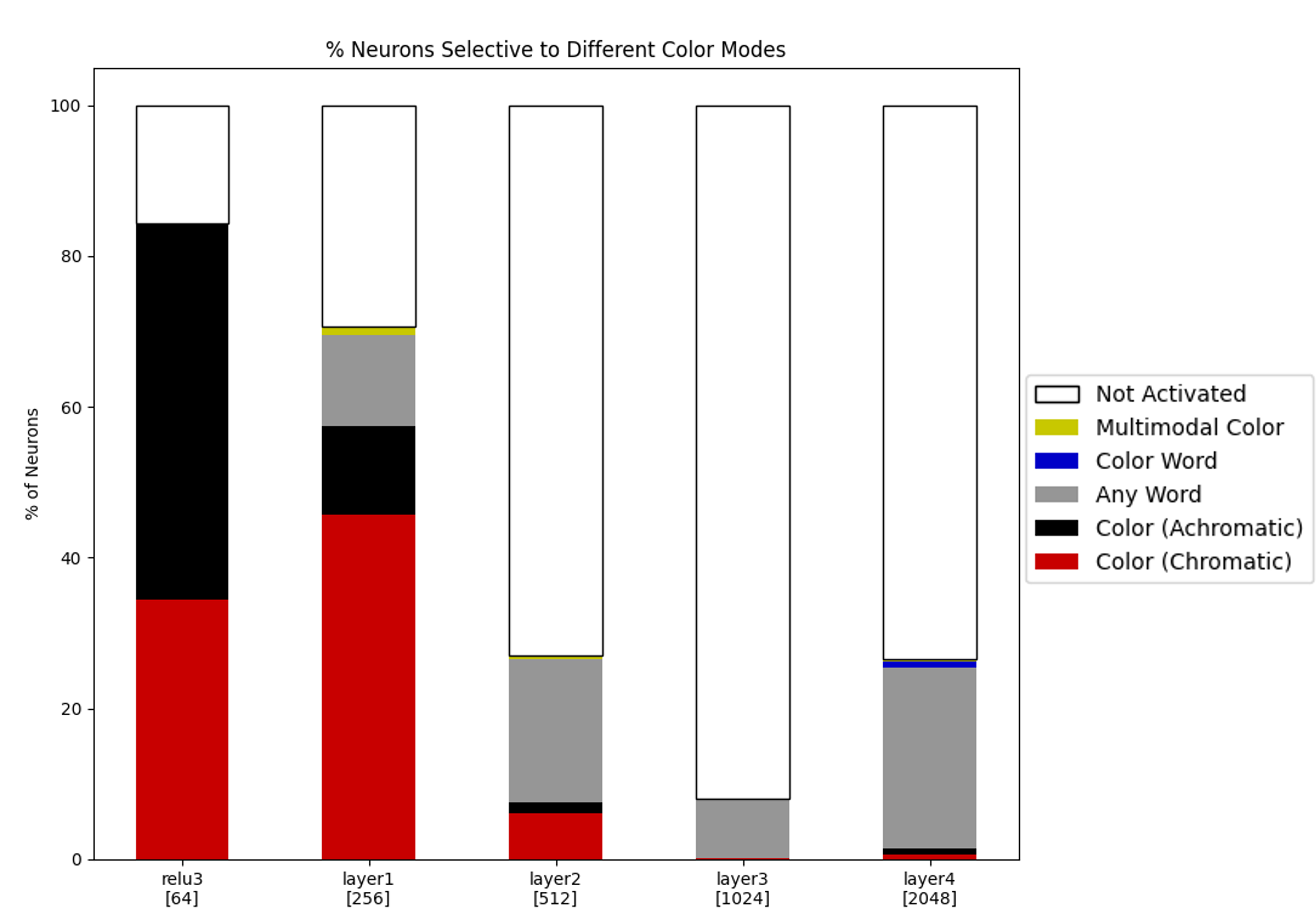}
\caption{Distribution of Neuron Types per layers in CLIP.}
\label{fig: neuron_modes}\vspace{-0.2cm}
\end{figure}

\noindent In figure \ref{fig: neuron_modes} we show the distribution of the different type of neurons, according to the previous description, that we found in CLIP layers. We can observe that in lower layers we have a big amount of achromatic color neurons whose activation underlies the representation of all the basic shapes and high frequencies of the text images. A second important group of color chromatic neurons which represent the basic color information. As we go deeper in the network the number of Color Word neurons are increasing, since more complex letters and words are being hierarchically built. One interesting finding is we have Color Multi-modal neurons in shallower layers, which is a novelty, since in previous works this kind of neurons has always been found  in deep layers and encoding high-level concepts. In this case, since our multi-modal neurons represent color which is a low-level property they are found earlier.  

\section{Conclusions}
In this work we explored how one of the most influential Visual-Language models in AI deals with color labelling tasks. We have performed a set of basic experiments to report how CLIP behaves in front of specific color tasks and we found out some that we summarized in next lines:
\vspace{0.25cm}
\begin{itemize}
\item \textbf{Achromatic stimuli are not related to the color concept.} It presents important errors when asked to assign black, white or grey labels. 
\item \textbf{Preference to label the  predominant color.} When asked for a global assignment, it labels the larger colored area, except when it is achromatic.
\item \textbf{Ability to attribute the color label to the asked image part.} It properly attributes the color label to object or background accordingly with the input text. Again with the exception of achromatic colors. If one of the parts is achromatic, the assigned label is always the chromatic one, independently of the  part asked in the input question.  
\item \textbf{Stroop effect with words written in white background.} It leans towards reading (80\%) rather than assigning the color of the font (16\%).

\item \textbf{Chomatic Backgrounds distract the reading preference in the Stroop test.} When distracted by a Chromatic background, CLIP still prioritizes reading (59\%), but with a  shift towards answering the color of the background (38\%), and completely ignoring the main objective, that is giving the color of the font (2\%).
\end{itemize}

\noindent Looking for an explanation to this behaviour is a hard task due the black-box nature of these models. We made some step toward this end. We analysed the internal representation at the neuron unit level. We developed a new Selectivity Index to identify neurons presenting a preference for specific types of labels. We have identified a set of Color Multi-Modal neurons, that combine its selectivity to written color words and the corresponding color stimulus. Interestingly, these neurons are found in shallow layers of the network, that could be due to the intrinsic nature of color as a generic attribute of any concept.

From the previous conclusions we hypothesize that the lack of understanding of achromatic stimuli as colors, could be due to their prevalence as image backgrounds in many datasets. A possible solution for a more robust and human-like minded model could be training the models more progressively, ensuring they learn from basic common-sense concepts to more complex ones, in a similar way as children learn about the world.

\section{Acknowledgments}
This work was supported by Grant PID2021-128178OB-I00 funded by MCIN/AEI/10.13039/501100011033 “ERDF/EU”

\begin{figure}[H]
\centering
\includegraphics[width=\linewidth]{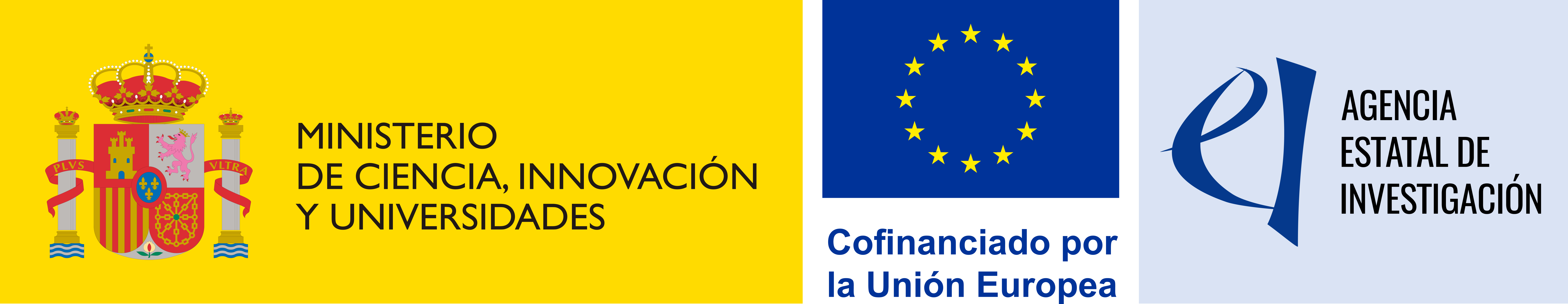}

\end{figure}

\bibliographystyle{plain}
\bibliography{biblio}


\end{document}